\documentclass{article}

\usepackage{arxiv}

\usepackage[utf8]{inputenc} 
\usepackage[T1]{fontenc}    
\usepackage{hyperref}       
\usepackage{url}            
\usepackage{booktabs}       
\usepackage{amsfonts}       
\usepackage{nicefrac}       
\usepackage{microtype}      
\usepackage{lipsum}
\usepackage{graphicx}
\usepackage{amsmath}
\usepackage{threeparttable}
\graphicspath{ {./images/} }

\title{CAST: Cross Attention based multimodal fusion of Structure and Text for materials property prediction}

\author{
    Jaewan Lee \\
  LG AI Research\\
  Seoul, Republic of Korea \\
  \texttt{jaewan.lee@lgresearch.ai} \\
   \And
    Changyoung Park\\
  LG AI Research\\
  Seoul, Republic of Korea \\
  \texttt{changyoung.park@lgresearch.ai} \\
  \And
  Hongjun Yang \\
  LG AI Research\\
  Seoul, Republic of Koreal \\
  \texttt{hongjun.yang@lgresearch.ai} \\
  \And
  Sungbin Lim \\
  Department of statistics\\
  Korea University\\
  Seoul, Republic of Korea \\
  \texttt{sunbin@korea.ac.kr} \\
  \And
  Woohyung Lim \\
  LG AI Research\\
  Seoul, Republic of Korea\\
  \texttt{w.lim@lgresearch.ai} \\
  \And
  Sehui Han \\
  LG AI Research\\
  Seoul, Republic of Korea\\
  \texttt{hansse.han@lgresearch.ai} \\
}

\begin{document}

\maketitle
\begin{abstract}
Recent advancements in graph neural networks (GNNs) have significantly enhanced the prediction of material properties by modeling crystal structures as graphs. However, GNNs often struggle to capture global structural characteristics, such as crystal systems, limiting their predictive performance. To overcome this issue, we propose CAST, a cross-attention-based multimodal model that integrates graph representations with textual descriptions of materials, effectively preserving critical structural and compositional information. Unlike previous approaches, such as CrysMMNet and MultiMat, which rely on aggregated material-level embeddings, CAST leverages cross-attention mechanisms to combine fine-grained graph node-level and text token-level features. Additionally, we introduce a masked node prediction pretraining strategy that further enhances the alignment between node and text embeddings. Our experimental results demonstrate that CAST outperforms existing baseline models across four key material properties—formation energy, band gap, bulk modulus, and shear modulus—with average relative MAE improvements ranging from 10.2 \% to 35.7 \%. Analysis of attention maps confirms the importance of pretraining in effectively aligning multimodal representations. This study underscores the potential of multimodal learning frameworks for developing more accurate and globally informed predictive models in materials science.

\end{abstract}

\section{Introduction}
With the advent of machine learning(ML) techniques, particularly graph neural networks(GNNs), researchers have been able to model material structures with unprecedented accuracy and speed by representing them as graphs and capturing intricate local interactions\cite{xie2018crystal, gasteiger2020directional, gasteiger2021gemnet, choudhary2021atomistic, ruff2024connectivity, merchant2023scaling}. Despite these advancements, the process of converting material structures into graph representations inherently leads to the loss of crucial information, such as crystal symmetries and the connectivity of repetitive structural units, which are critical to certain material properties. In addition, due to the limitations of GNN architectures, Gong et al \cite{gong2023examining} found that recent state-of-the-art GNNs do not capture the periodicity of crystal structures well.
To address this limitation, multimodal learning can be leveraged to complement the lost information by incorporating additional modalities. While multimodal learning has been extensively studied in fields such as vision, language, and speech \cite{wang2023large, xu2023multimodal, park2022graph}, its application in the materials, especially inorganic crystals, remains relatively underexplored.\cite{das2023crysmmnet, munjal2024lattice, ozawa2024graph, ock2024unimat, lee2023clcs}
A key factor behind this gap was the scarcity of appropriate multimodal datasets in materials science. With recent advancements, such as robocrystallographer\cite{ganose2019robocrystallographer}—an API that generates textual descriptions from crystal structures—have opened new avenues for integrating multimodal approaches in materials science research. It generates text descriptions based on rules derived from structural information, including global features that are often lost when structures are converted into graph representations.

Current multimodal approaches in materials science primarily rely on techniques, such as embedding concatenation or contrastive pretraining methods exemplified by CLIP\cite{radford2021learning}. Notably, CrysMMNet\cite{das2023crysmmnet} and MultiMat \cite{moro2025multimodal} have employed these methods with success. CrysMMNet encodes structural and textual information separately before concatenating their embeddings, whereas MultiMat derives embeddings from structural data, textual information, density of states (DOS), and charge density prior to contrastive learning. Both approaches achieve better performance over unimodal GNNs. Nevertheless, two important questions remain unaddressed. 
First, coarse-grained(material-level) modality fusion approaches, used in CrysMMNet and MultiMat, may not fully exploit the intricate inter-modal relationships, potentially limiting model performance. Second previous studies have not adequately justified the use of a text encoder instead of directly utilizing the structural descriptors employed by robocrystallographer.

In this paper, we proposed a CAST, Cross Attention-based multimodal fusion of Structure and Text, which integrates graph and text modalities at a fine-grained level(atom-level), ensuring profound mutual interactions and conducted an ablation study to verify a benefit of using a text encoder. Our method incorporates atom- and token-level features using cross-attention mechanisms, which are proposed in a transformer\cite{vaswani2017attention}, allowing the model to learn more complex relationships between the structural and textual representations. 
Inspired by the success of masked object prediction in the language domain \cite{radford2018improving, kenton2019bert, liu2019roberta}, we applied a Masked Node Prediction(MNP) pretraining strategy by masking a subset of nodes in the graph and training the model to predict the masked nodes using information from their neighboring nodes and corresponding text tokens. 
Through extensive experimentation, we demonstrated that our approach outperforms unimodal and multimodal baselines(e.g., CrysMMNet and MultiMat), achieving average relative improvements ranging from 10.2\% to 35.7\% across four property prediction tasks.
Analyses of cross-attention maps further revealed that pretraining enables attention heads to effectively capture a diverse range of node-token relationships, which are critical for accurate predictions. By addressing the inherent limitations of existing models, our work establishes a robust framework for multimodal learning in materials science, paving the way for more accurate predictive models. 
\section{Results and Discussion}

\subsection{The overview of CAST Framework} 
\begin{figure}
  \centering
  \includegraphics[width=\textwidth]{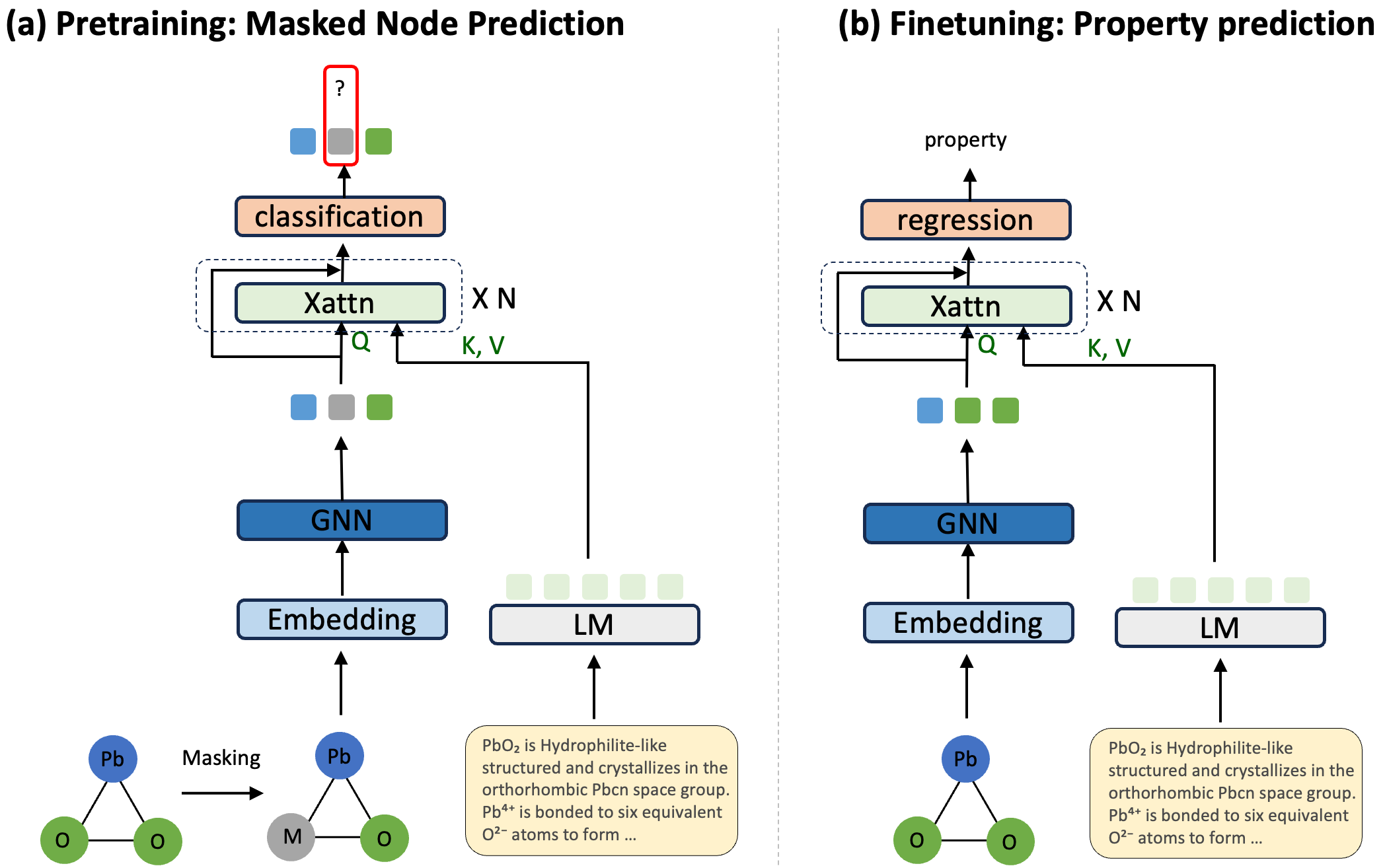}
  \caption{\textbf{Overview of the CAST framework.}
 \textbf{(a)Pretraining:} Mask 50\% of graph nodes and predict their element types via cross attention between node embeddings and text embeddings.
 \textbf{(b)Finetuning:} Replace classification head with a regression head to predict material properties using the aligned multimodal embeddings.}

  \label{fig:figure1}
\end{figure}

The schematic of CAST is illustrated in Figure \ref{fig:figure1}. The framework integrates structural and textual data through two stages: (1) masked node prediction (MNP) pretraining for multimodal alignment and (2) finetuning for material property prediction. Initially, structural and textual modalities are embedded separately using a structure encoder (coGN\cite{ruff2024connectivity}) and a text encoder (MatSciBERT\cite{gupta2022matscibert}). These embeddings are then fused at a fine-grained level through a cross-attention mechanism inspired by transformer architectures, enabling detailed interactions between node embeddings from structural graphs and token embeddings from textual descriptions. During the MNP pretraining stage, nodes in the structural graph are partially masked to encourage the model to leverage textual information effectively, thereby aligning the structural nodes closely with relevant text tokens. Subsequently, during the finetuning stage, the pretrained cross-modal embeddings are utilized for precise material property prediction, substantially enhancing performance through improved multimodal representation.

\subsection{Comparison of Predictive Performance}
The CAST was compared against unimodal models(coGN and MatSciBERT) and other multimodal methods(CrysMMNet and  MultiMat), across four material properties: total energy($E_{tot}$), bandgap, shear modulus($G_{vrh}$), and bulk modulus($K_{vrh}$). To evaluate the effect of MNP pretraining, we define two settings: CAST, which is pretrained on MNP and then fine-tuned, and CAST-base, which is trained directly on the regression tasks without pretraining. The results were summarized in Table \ref{tab:model evaluation}. CrysMMNet and MultiMat utilized different encoders for their respective architectures. To ensure a fair comparison of fusion methods, we standardized the encoders by using coGN as the structure encoder and MatSciBERT as the text encoder for training both models. In MultiMat, contrastive learning is originally conducted using four modalities, but in this study, we used only structure and text for training. While CrysMMNet originally froze the text encoder during training, we evaluated both the frozen approach(CrysMMNet) and a fine-tuning approach using LoRA \cite{hu2021lora}(CrysMMNet-LoRA), a widely adopted technique for efficient model fine-tuning. Since the MatSciBERT was not pretrained for regression tasks, incorporating LoRA allowed us to explore and compare a broader range of multimodal scenarios within the concatenation framework. Details on the implementation of each model are provided in the Methods section.

To quantify the performance gain of our method, we computed the average relative improvement of CAST over each baseline across the four property prediction tasks. 
Specifically, CAST achieved average relative MAE improvements compared to coGN(28.8\%), MatSciBERT(24.7\%), CrysMMNet(24.6\%), CrysMMNet-LoRA(10.2\%), MultiMat (35.7\%), and CAST-base(7.3\%). These results highlight the effectiveness of our model in producing consistently more accurate predictions.

Comparisons of single-modality models indicated distinct strengths and weaknesses; the text-based MatSciBERT demonstrated superior performance in predicting total energy, whereas coGN excelled in bandgap prediction. For bulk and shear modulus predictions, the two models performed similarly. These complementary strengths across different properties demonstrated the promise of multimodal approaches, suggesting that integrating text and graph modalities could harness their unique advantages for more robust material property prediction. Building on these insights, CrysMMNet outperformed coGN on three properties—band gap, shear and bulk modulus—but was still outperformed by MatSciBERT on total energy. Fine-tuning CrysMMNet with LoRA (CrysMMNet–LoRA) further reduced MAE in formation energy, shear and bulk modulus, although gains in band-gap prediction remained modest. While these results demonstrated the potential of multimodal fusion, the fact that it does not consistently outperform unimodal models across all four properties underscores the limitations of coarse-grained multimodal approaches.

The performance of MultiMat provided two lessons. First, unlike the findings reported in the MultiMat study, our results showed that contrastive learning alone during pretraining led to lower performance compared to the unimodal coGN model. Further research is needed to clarify whether this degradation arises from dataset characteristics or the choice of encoders. Second, the superior performance of a concatenation approach over contrastive learning suggested that directly utilizing both modalities during fine-tuning could enhance predictive capabilities. Therefore, future multimodal models should explicitly integrate modalities throughout both pretraining and fine-tuning stages to fully exploit complementary multimodal information.

CAST not only explicitly integrates modalities but also leverages fine-grained information, resulting in significant performance gains. Notably, even without pretraining, the CAST-base model achieved an average 3.2\% reduction in MAE compared to the best baseline, CrysMMNet-LoRA. When pretraining was applied, CAST further improved performance reducing MAE on three of the four properties and trailing the best baseline by only 0.001 MAE on log(Kvrh), thereby demonstrating its overall effectiveness. An additional ablation study on node masking ratios (Supplementary Information 5.1) showed that a higher masking ratio (50\%) improved prediction performance by an average of 3.9\% across four properties, indicating that a more challenging node prediction task promotes better representation alignment and enhanced predictive outcomes. Specifically, for bandgap prediction, most multimodal models, including CAST-base, underperformed compared to the unimodal coGN. However, pretraining alleviated this issue, leading to a 7.1\% performance improvement over coGN. This result underscores the critical importance of pretraining in multimodal learning.

\begin{table}
\centering
\resizebox{\textwidth}{!}{
  \begin{tabular}[width=\textwidth]{c c c c c}
    \toprule
    Models  & $E_{tot}$ &  $Bandgap$ & $log(G_{vrh})$  & $log(K_{vrh})$  \\
    \midrule
    coGN & 0.673(0.118) & 0.381(0.006) & 0.091(0.006) & 0.050(0.001)\\
    MatSciBERT & 0.390(0.047) & 0.420(0.002) & 0.089(0.002) & 0.053(0.001) \\
    \midrule
    CrysMMNet & 0.615(0.121) & \underline{0.369(0.005)} & 0.085(0.003) & 0.047(0.001) \\
    CrysMMNet-LoRA & 0.332(0.056) & 0.416(0.005) & 0.073(0.004) &  \textbf{0.038(0.001)}\\
    MultiMat & 1.095(0.086) & 0.414(0.003) & 0.089(0.002) & 0.055(0.002) \\
    \midrule
    CAST-base & \underline{0.277(0.051)} & 0.394(0.003) & \underline{0.072(0.002)} & 0.042(0.002)\\
    \textbf{CAST} & \textbf{0.256(0.045)} & \textbf{0.354(0.002)} & \textbf{0.069(0.001)} & \underline{0.039(0.0003)}\\
    \bottomrule
\end{tabular}
}
\caption{\textbf{Prediction performance comparison table} 
Comparison of predictive performance (MAE) of seven models over three random seeds; mean values are shown with standard deviations in parentheses. Best results are bolded and second-best are underlined.}
\label{tab:model evaluation}
\end{table}

\subsection{Further analysis of pretraining effects}
\begin{figure}
  \centering
  \includegraphics[width=\textwidth]{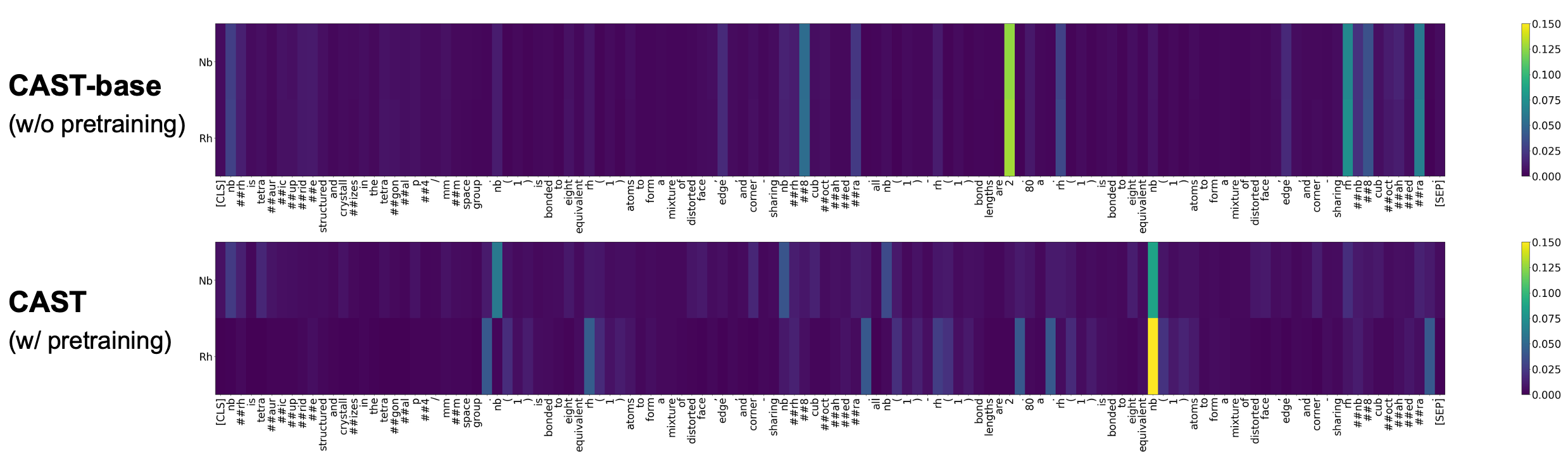}
  \caption{\textbf{Cross attention maps for test set crystal mp-1963.} \textbf{Top:} CAST-base(w/o pretraining) showed uniform attention(stripe patterns) \textbf{Bottom:} CAST(w/ pretraining) exhibited more diverse attention across tokens, indicating stronger cross modal alignments. 
  More examples can be seen in supplementary information Figure S5.}
  \label{fig:figure2}
\end{figure}

We conducted an analysis to investigate how pretraining contributed to the observed improvement in predictive performance. Using a test set example, we examined the attention maps generated by the cross-attention mechanism (Figure \ref{fig:figure2}). In the figure, the x-axis represents the text tokens acting as keys, while the y-axis corresponds to the query tokens. Each cell indicates the attention score, with brighter colors reflecting higher scores and stronger associations.

The first row shows the attention maps extracted from CAST-base, whereas the second row represents the attention maps obtained from CAST. In the CAST-base case, we observed that all nodes predominantly attended to the same text tokens, resulting in stripe-like attention patterns. 
In contrast, CAST demonstrated a distinctly diverse attention distribution, with each node attending differently to a wider variety of text tokens. This indicates that pretraining facilitates the model's ability to form associations that enable nodes to attend more broadly and distinctly to various text tokens. Additional examples demonstrating this behavior are provided in supplementary Figure S5.

\begin{figure} 
  \centering
  \includegraphics[width=\textwidth]{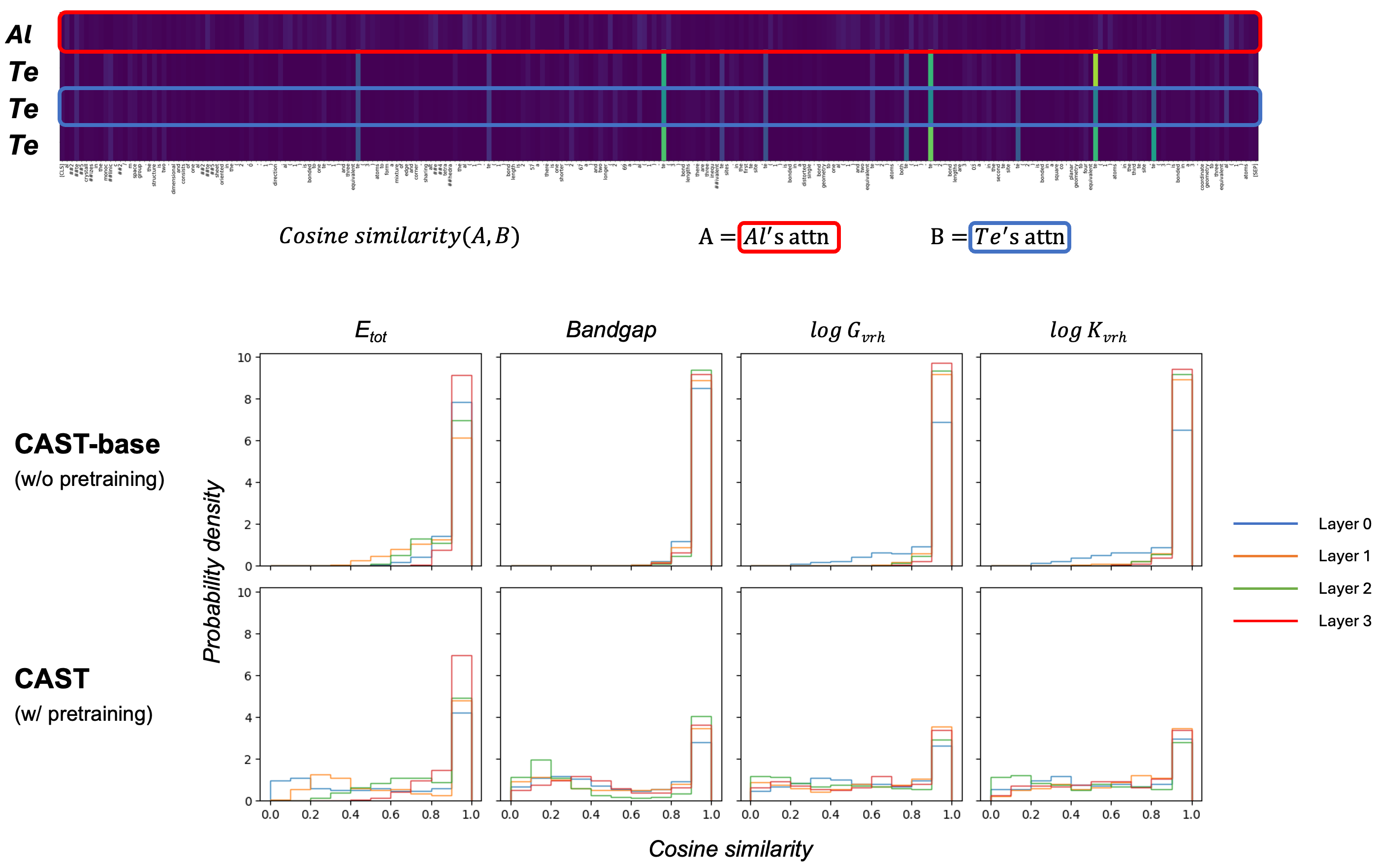}
  \caption{\textbf{Distributions of cosine similarities between node pair attention vectors across layers.}
   An example illustrated the computation of cosine similarity between pairs of node attention vectors. Top row showed the distribution of cosine similarity values for CAST-base(w/o pretraining), with most values concentrated near 1. Bottom row showed the distribution for CAST(w/ pretraining), exhibiting a broader range that reflects more diverse node–token alignments.
  }
  \label{fig:figure3}
\end{figure}

To quantitatively analyze these observations, we plotted the probability density distribution of the similarities in the attention maps between node tokens across the testset.(Figure \ref{fig:figure3}). 
In the given example, the material is represented as a graph with four nodes (Al, Te, Te, Te). To quantify how diversely node tokens reference text tokens, we calculated the cosine similarity of attention values between all pairs of nodes (combinations), as defined in Equation~\ref{eq:cosine_similarity}. In Equation~\ref{eq:cosine_similarity}, $\mathbf{A}$ and $\mathbf{B}$ represent the attention value vectors of two different node tokens over the same text tokens. The index $i$ runs over all text tokens, and $n$ denotes the total number of text tokens.

\begin{equation}
\text{cosine similarity}(\mathbf{A}, \mathbf{B}) = 
\frac{\mathbf{A} \cdot \mathbf{B}}{\|\mathbf{A}\| \, \|\mathbf{B}\|} = 
\frac{\sum_{i=1}^{n} A_i B_i}{\sqrt{\sum_{i=1}^{n} A_i^2} \, \sqrt{\sum_{i=1}^{n} B_i^2}}
\label{eq:cosine_similarity}
\end{equation}

This analysis offers a quantitative measure of how pretraining enhances the diversity of text token references by node tokens.
In the CAST(w/o pretraining), the similarities were predominantly concentrated near \texttt{1}, indicating a high degree of focus that intensified as layers progressed. On the other hand, the pretrained model exhibited a wider distribution of similarities, with less severe concentration as layers deepened compared to the CAST(w/o pretraining).

These findings demonstrate that pretraining enables node tokens to reference a diverse array of text tokens, potentially enhancing their capacity to capture critical and relevant information. This observation suggests that the phenomenon illustrated in Figure \ref{fig:figure2} is not an isolated occurrence but represents a consistent trend across the entire dataset. While the ability of node tokens to attend diversely and effectively extract essential information from a wide range of relevant text tokens highlights the model's capability in capturing intricate relationships between modalities, further evidence is needed to conclusively determine the impact on downstream performance.

\subsection{Ablation: Feasibility analysis of using a text encoder}

\begin{figure}
  \centering
  \includegraphics[width=\textwidth]{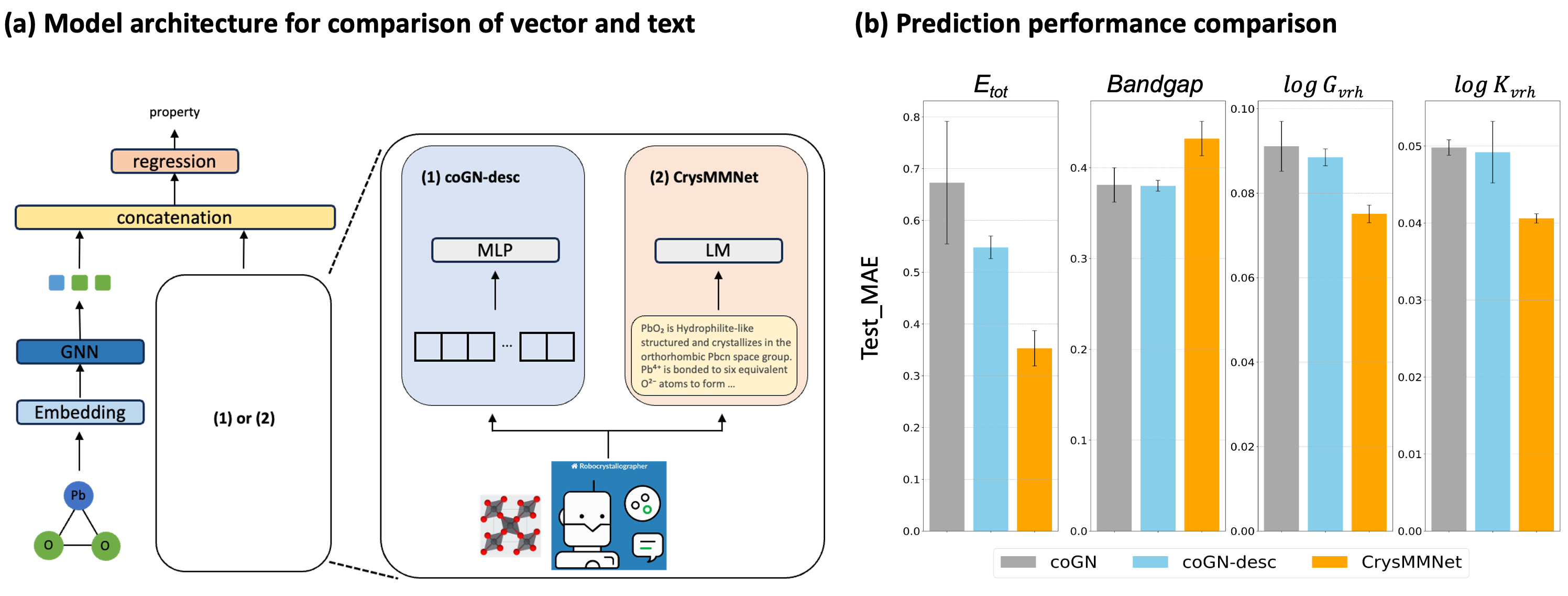}
  \caption{\textbf{Descriptor vs text embeddings.}
  \textbf{(a)  Model architecture for comparison of vector and text} coGN-desc concatenated descriptor vectors with GNN embeddings. CrysMMNet concatenates text embeddings.
  \textbf{(b) Prediction performance comparison} coGN-desc achieved an average test MAE reduction of 5.7 \% across four properties. CrysMMNet achieved an average reduction of 17.5 \%, thereby outperforming coGN-desc by 13.6 \%.
  }
  \label{fig:figure0}
\end{figure}

To assess whether incorporating global information in multimodal learning necessarily requires the use of a text encoder, we conducted experiments comparing it with a simpler alternative—using descriptors that compactly represent global information. As shown in Figure~\ref{fig:figure0}, we evaluated which approach yields better performance. Further details are provided in the Methods section. 

We compared the mean absolute error (MAE) on the test set for predicting total energy, bandgap, and the logarithms of shear modulus and bulk modulus across three scenarios, as depicted in Figure 4(b): (1) coGN alone, (2) coGN-desc, which concatenates descriptor vectors with graph embeddings, and (3) CrysMMNet, which concatenates text embeddings derived from a language model with graph embeddings.
Our findings indicate that integrating global information improved performance by an average of 5.7\% using descriptor vectors (coGN-desc) and 17.5\% using text embeddings (CrysMMNet) compared to coGN alone across four properties. Additionally, embeddings derived from the language model significantly outperformed descriptor-based vectors, achieving an average improvement of 13.6\% over the descriptor-based approach. Thus, our experiments validate that complementing graph embeddings with text embeddings from a language model is more effective.
Two primary reasons may explain this result. First, the language model was pretrained on a large corpus directly related to crystals, enabling more effective embedding of relevant semantic information.
Second, in conventional vector representations, categorical variables do not inherently capture meaningful differences numerically. In contrast, text embeddings clearly distinguish between distinct categories, thereby minimizing the information loss that typically occurs during vectorization.
Third, for numerical variables, encoding not only the values but also their corresponding variable names further reduces information loss when using a language model. A similar phenomenon was observed in the TransTab study \cite{wang2022transtab}, which encoded both cell values and their associated column headers, thereby incorporating semantic context and enhancing the model’s ability to interpret tabular data.

\section{Conclusion}
In this study, we introduced CAST(Cross Attention-based multimodal fusion of Structure and Text), an innovative framework for material property prediction. By integrating structural and textual data at a fine-grained level through cross attention mechanisms and combining it with a masked node prediction (MNP) pretraining strategy, CAST achieved outstanding performance—outperforming unimodal models (coGN and MatSciBERT) and other multimodal approaches (Concatenation and Contrastive learning) by 10.2–35.7\% on average MAE across four key properties: total energy, bandgap, and bulk/shear modulus. This synergy between architectural design and pretraining not only improved predictive accuracy but also enhanced model robustness, particularly for the challenging property of bandgap.

The MNP pretraining strategy was pivotal in unlocking the full capability of the CAST framework, yielding an additional 7.3\% average improvement compared to CAST without pretraining(CAST-base). This strategy potentially improved the alignment between structural node tokens and corresponding textual tokens, fostering richer and more interpretable attention patterns. Such alignment contributed to more expressive multimodal representations and highlights the value of tailored pretraining techniques in enhancing model performance. Additionally, we demonstrated that leveraging a sophisticated language model to incorporate global textual information with graph representations significantly outperforms traditional descriptor-based methods by 13.6\%, underscoring the advantages of using a text encoder for capturing comprehensive material characteristics.

Nevertheless, CAST faces scalability challenges due to the substantial computational demands of sophisticated language models and attention mechanisms, particularly when applied to large or complex datasets. In addition, the current limitation of the text encoder to 512 tokens restricts its ability to exploit longer and potentially more informative textual descriptions. Addressing these issues—by exploring more efficient fusion methods, incorporating lightweight or advanced encoders, and extending the framework to a broader range of multimodal datasets—will be critical for enhancing the utility of CAST. Moreover, in the MNP task, element-level masking was employed to match atomic types, which imposes a limitation in distinguishing atoms of the same type but with different local environments. Future work will aim to overcome these constraints to further improve the model’s generalizability and applicability.

\section{Methods}

\subsection{Data}
\subsubsection{Preparation}
For training and evaluation, data was downloaded from the Materials Project\cite{jain2013commentary} database and rigorously filtered to ensure quality and relevance. The filtering criteria were inspired by the data cleaning methods used in MatBench\cite{dunn2020benchmarking} and further enhanced with additional constraints to ensure practical applicability and facilitate efficient screening processes for real-world use. The applied filters are as follows:
\newline

\noindent\textbf{MatBench filtering criteria:} 
\begin{list}{$\bullet$}{\setlength{\leftmargin}{2em}}
    \item Remove entries with a formation energy or energy above the convex hull greater than 150 meV/atom.
    \item Exclude entries where $G_{\text{Voigt}}, G_{\text{Reuss}}, G_{\text{VRH}}, K_{\text{Voigt}}, K_{\text{Reuss}},$ or $K_{\text{VRH}}$ are less than or equal to zero.
    \item Remove entries that do not satisfy the conditions $G_{\text{Reuss}} < G_{\text{VRH}} < G_{\text{Voigt}}$ or $K_{\text{Reuss}} < K_{\text{VRH}} < K_{\text{Voigt}}$.
    \item Exclude entries containing noble gases.

\end{list}
\vspace{\baselineskip}
\noindent \textbf{Additional filtering:}
\begin{list}{$\bullet$}{\setlength{\leftmargin}{2em}}
    \item Exclude entries with a formation energy lower than -10 eV/atom.
    \item Remove entries with shear modulus or bulk modulus values exceeding 1000 GPa.
\end{list}
\vspace{\baselineskip}
\noindent For regression tasks, we used four properties($E_{tot}$, bandgap, $log(G_{vrh})$ and $log(K_{vrh})$). Each dataset comprised 114.3k, 65.4k, 9.4k and 9.4k instances, respectively.

Detailed statistical analyses of the regression data for each target property can be found in the Table \ref{tab:statistic_analysis}. For the pretraining stage, we leveraged the $E_{tot}$ data, which had the largest volume of samples.

\subsubsection{Text generation}
While the Materials Project directly provides structural information in cif formats, we generated corresponding textual descriptions with the Robocrystallographer API using its default settings.  In the rare cases (less than 0.2\%) where no text was produced, we simply passed the [CLS] token through the text encoder. These instances were negligible and unlikely to affect our results. Statistical details—including the number of tokens per description and the success rate of text generation—are provided in Table \ref{tab:statistic_analysis}.

\subsubsection{Statistical analysis of data}
The Table \ref{tab:statistic_analysis} presents a comprehensive summary of the statistics of the data set for $E_{tot}$, bandgap, the $log(G_{vrh})$ and the $log(K_{vrh})$, with the data partitioned into training, validation, and test sets in a ratio of 8:1:1.
For each property, key characteristics of the structure graphs and texts are detailed, such as the mean and standard deviation of the number of graph nodes and text tokens. The text existence rate indicates the proportion of samples with successfully generated textual descriptions and absolute counts are provided below. Furthermore, the table reports the mean and standard deviation of the target property values, offering insights into the distributions within the dataset. These statistics indicate the variation in structural and textual complexities across properties and highlight the near-complete availability of texts, enabling the seamless integration of multimodal information for predictive modeling tasks.

\begin{table}[htbp]
\scriptsize
\centering
\resizebox{\textwidth}{!}{
    \begin{tabular}{|l|l|l|l|l|l|l|l|l|}
    \hline
     & \multicolumn{2}{c|}{\textbf{$E_{tot}$}} & \multicolumn{2}{c|}{\textbf{$Bandgap$}} & \multicolumn{2}{c|}{\textbf{$log(G_{vrh})$}} & \multicolumn{2}{c|}{\textbf{$log(K_{vrh})$}} \\ \hline
     \textbf{Data type} & \textbf{train} & \textbf{val/test} & \textbf{train} & \textbf{val/test} & \textbf{train} & \textbf{val/test} & \textbf{train} & \textbf{val/test} \\ \hline 
    Dataset size & 91,520 & 22,878 & 52,371 & 12,996 & 7,513 & 1,910 & 7,513 & 1,910 \\ \hline
    \# of node mean & 17.28 & 17.10 & 22.95 & 22.78  & 3.67 & 3.58 & 3.67 & 3.58\\ \hline
    \# of node std & 28.59 & 28.34 & 34.95 & 34.76  &2.03 &2.39 &2.03 &2.39\\ \hline
    \# of tokens mean & 1,796.65 & 1,783.56 & 2,326.0 & 2,323.2 & 250.83 & 250.16 & 250.83 & 250.16 \\ \hline
    \# of tokens std & 3,574.83 & 3,516.19 & 4,137.72 & 4,109.76 & 272.62 & 327.55 & 272.62 & 327.55 \\ \hline
    Text existence rate & 99.9\% & 99.9\% & 99.9\% & 99.8\%& 100\% & 100\% & 100\% & 100\% \\ \hline
    \# of existing text & 91,437 & 22,855 & 52,298 & 12,974& 7,513 & 1,910 & 7,513 & 1,910 \\ \hline
    Y mean & -9.14 & -9.26 & 2.17 & 2.15 & 1.54 & 1.55 & 1.87 & 1.88 \\ \hline
    Y std & 7.22 & 7.38 & 1.54 & 1.55 & 0.39 & 0.38 & 0.38 & 0.37  \\ \hline
\end{tabular}
}
\caption{\textbf{Statistical analysis of data} The table provides a comprehensive summary of the dataset statistics across different material properties, including $E_{tot}$, $bandgap$, $log(G_{vrh})$, and $log(K_{vrh})$, as well as their respective data splits for train and val/test.}
\label{tab:statistic_analysis}
\end{table}

\subsubsection{Characteristics of Text and Descriptors}

To analyze the characteristics of text and descriptor-based representations in greater detail, it is important to highlight their structural differences. Text is inherently unstructured data, meaning it can be directly utilized within the model without requiring additional preprocessing, particularly when leveraging pre-trained language models. In contrast, descriptors are structured data, comprising numerical, categorical, and boolean types, which necessitate a digital transformation process before being fed into the model. A concrete example of these differences is shown below and in Fig S1.

\begin{itemize}
    \item \textbf{Categorical type}: In text, the space group is described by phrases such as \textit{``Pbcn space group''}, while in descriptor vectors, it is represented numerically as \texttt{0, 1, 2} and so on.
    \item \textbf{Numerical type}: 
    In text, bond lengths are described by phrases such as \textit{``bond distances ranging from 2.14 to 2.24 angstroms''}, whereas in feature vectors, they are represented numerically as \texttt{2.14} and \texttt{2.24}.
    \item \textbf{Boolean type}: 
    In text, edge-sharing connectivity appears as phrases like \textit{``forms an edge-sharing PbO\textsubscript{6} octahedra''}, whereas in feature vectors, this is represented numerically as \texttt{1}.
\end{itemize}

Distribution of robocrystallographer descriptors is illustrated in supplementary information section 3.

\subsection{Encoders for each modality}
\subsubsection{Structure Encoder : coGN}
By leveraging the inherent symmetry of crystals, coGN employs an asymmetric unit cell representation to reduce the number of graph nodes, thereby improving computational efficiency. The proposed Nested Line Graph Network (NLGN) architecture further enhances the GNN framework through optimized message passing. This approach has demonstrated superior performance across most tasks within the MatBench benchmark dataset. Key factors contributing to these performance improvements include optimized connectivity and enriched message-passing capabilities. Given its state-of-the-art (SOTA) performance on MatBench and computational efficiency, coGN is considered to have robust encoding capabilities, making it an ideal choice as the GNN encoder for our work.

\subsubsection{Text Encoder: MatSciBERT}
MatSciBERT is a domain-specific language model designed for materials science, trained on a large corpus of peer-reviewed materials science publications. It excels at information extraction tasks by effectively interpreting the unique notations and terminology prevalent in materials science literature. Evaluations on tasks related to materials such as abstract classification, named entity recognition, and relation extraction have demonstrated that MatSciBERT outperforms general scientific language models, such as SciBERT\cite{Beltagy2019SciBERT}. The pretrained and fine-tuned models are publicly available, serving as a valuable resource for the materials science community. We hypothesize that leveraging a model with a deep understanding of materials science will significantly enhance predictive performance. Therefore, we adopted MatSciBERT as the text encoder in our framework.

\subsection{Multimodal fusion methods}
We compared regression performance using three approaches: CAST, CrysMMNet, MultiMat and coGN-Desc. To ensure a fair comparison, we standardized the training hyperparameters across all methods. For training, we employed a periodic cosine scheduling strategy to adjust the learning rate. To enhance training stability, we included a warm-up phase for the initial 1,000 steps. For contrastive learning pretraining, we set the batch size to 360, following the configuration used in the MultiMat paper, as larger batch sizes are more efficient for this approach. For other pretraining task and regression tasks, a batch size of 64 was used. For consistency, we employed coGN and MatSciBERT as the structure and text encoders across all methods. Accordingly, the feature dimensions of the node tokens and text tokens were set to 128 and 768, respectively. 

\subsubsection{CAST(proposed method)}
CAST, based on cross-attention, followed a two-step process: pretraining and finetuning. Unlike previous studies \cite{das2023crysmmnet, moro2025multimodal, munjal2024lattice} in the materials domain that focused on multimodal fusion at the instance level, we hypothesized that token-level fusion can achieve superior performance. To enable interaction at the token level, we employ the cross attention mechanism of the transformer. During the pretraining, a subset of graph nodes was randomly masked with a 50\% probability. The structure encoder processed the graph to generate embeddings for each node, while the corresponding text was passed through a text encoder to produce token-level embeddings. These embeddings interacted through a cross-attention mechanism, where the node embeddings served as queries, and the text token embeddings acted as keys and values. This interaction allowed the node embeddings to incorporate contextual information from the text. As the model learned to predict the element types of the masked nodes, it simultaneously aligned node embeddings with the most relevant text tokens. The fusion module consisted of four cross attention layers with eight attention heads, each with an attention dimension of 128. The classification block is a single linear layer. Following the pretraining phase, the model underwent a finetuning stage specifically designed for property prediction tasks. In this phase, the pretrained model was preserved, but the classification block was replaced with a regression block, also a single linear layer, to align with the regression objective. By leveraging the structural and textual embeddings aligned during pretraining, the model was optimized to predict material properties with improved accuracy. This gain reflected the enhanced cross-modal understanding that enable seamless integration of graph and text features. As shown in Table \ref{tab:model evaluation}, our method demonstrated more robust and precise performance compared to other unimodal and multimodal models.

\subsubsection{CrysMMNet}
CrysMMNet, to our knowledge, is the first approach to leverage multimodal learning to predict material properties. Their model employs ALIGNN as the structure encoder and MatSciBERT as the text encoder. In this framework, the graph structure was processed through a graph encoder to generate graph embeddings, while textual descriptions were passed through a text encoder and a projection layer to produce text embeddings. These representations were then fused to jointly model the input modalities and predict crystal properties. However, their approach froze the text encoder during training and did not assess how an unfrozen encoder affects performance. To fill this gap, we conducted additional experiments, comparing the performance of the concatenation method with a frozen language model(LM) and LoRA fine-tuning. The results of these experiments are summarized in Table \ref{tab:model evaluation}.

\subsubsection{MultiMat}
In MultiMat, contrastive learning between multimodal representations was utilized to pretrain the structure encoder, similar to the approach used in CLIP. Their study reported that applying the pretrained graph encoder to downstream tasks improved performance compared to training from scratch. MultiMat leveraged four modalities: crystal structures, density of states (DOS), and charge density, while our study focused on comparing different modality fusion strategies.
To ensure a fair comparison with other methods, we pretrained the model using only structures and texts. Furthermore, while MultiMat utilizes PotNet \cite{lin2023efficient} as the structure encoder and MatBERT \cite{walker2021impact} as the text encoder, we adapted the model by employing the coGN structure encoder and MatSciBERT text encoder. This alignment with the configurations of other methods ensures fair evaluation across models.

\subsubsection{coGN-desc}
In coGN-desc, we encoded material descriptors as numeric vectors rather than text tokens and concatenate them with graph embeddings. Our architecture largely followed Gong et al.\cite{gong2023examining}, differing only in this descriptor input module. Of the original 157 descriptors, the “corner sharing octahedral tilt angle” was excluded due to 77,931 \texttt{NaN} entries. We encoded categorical descriptors using scikit-learn’s OrdinalEncoder\cite{scikit-learn}—noting that this approach imposed an arbitrary order on categories-and used numerical and boolean descriptors directly. To evaluate representation strategies, we compared concatenating GNN embeddings with global text embeddings versus descriptor vectors; the setup is illustrated in Figure \ref{fig:figure0}(a). Descriptor vectors were projected through a single linear layer followed by a SiLU activation\cite{hendrycks2016gaussian}.

\section*{Data availability}
Access to the Materials Project\cite{jain2013commentary} is available at https://next-gen.materialsproject.org/ or Materials Project API.

\section*{Acknowledgement}
This work was supported by LG AI Research.

\section*{Author contributions}
J.L. implemented the algorithms, code and analyses described in this work and wrote the manuscript. C.P. discussed the results and wrote the manuscript. Other authors discussed the results and gave advice on the manuscript.

\section*{Competing interests}
The authors declare no competing interests.

-----------------

\bibliographystyle{unsrt}  
\bibliography{references}  






\end{document}